\documentclass{article}
\usepackage{aaai19}
\usepackage{times}  
\usepackage[noend]{algorithmic}
\usepackage{amsmath}
\usepackage{amsthm}
\usepackage{amsbsy}
\usepackage{amsfonts}
\usepackage[ruled,linesnumbered]{algorithm2e}
\usepackage{multirow}
\usepackage{graphicx}
\usepackage{epsfig}
\usepackage{epstopdf}
\usepackage{amssymb}
\usepackage{subfigure}
\usepackage{wrapfig}
\usepackage{float}
\usepackage{color}
\usepackage{url}
\newtheorem{proposition}{Proposition}

\newtheorem{definition}{Definition}

\newenvironment{proofsketch}{%
	\proof}{\endproof}
\newcommand{\zo}{0/1\,RM}

\title{Minimizing the Negative Side Effects of Planning with Reduced Models}
\author{Sandhya Saisubramanian \and Shlomo Zilberstein\\[4pt]
	College of Information and Computer Sciences \\
	University of Massachusetts\\
	Amherst, MA 01003, USA \\
\{saisubramanian,shlomo\}@cs.umass.edu   }

\nocopyright
\begin{document}
		
\maketitle
\begin{abstract}
Reduced models of large Markov decision processes accelerate planning by considering a subset of outcomes for each state-action pair. This reduction in reachable states leads to replanning when the agent encounters states without a precomputed action during plan execution. However, not all states are suitable for replanning. In the worst case, the agent may not be able to reach the goal from the newly encountered state.  Agents should be better prepared to handle such risky situations and avoid replanning in risky states.  Hence, we consider replanning in states that are unsafe for deliberation as a \emph{negative side effect} of planning with reduced models. While the negative side effects can be minimized by always using the full model, this defeats the purpose of using reduced models. The challenge is to plan with reduced models, but somehow account for the possibility of encountering risky situations. An agent should thus only replan in states that the user has approved as safe for replanning. To that end, we propose planning using a portfolio of reduced models, a planning paradigm that minimizes the negative side effects of planning using reduced models by alternating between different outcome selection approaches. We empirically demonstrate the effectiveness of our approach on three domains: an electric vehicle charging domain using real-world data from a university campus and two benchmark planning problems.
\end{abstract}

\begin{figure}[t]
	\centering
	\includegraphics[scale=0.35]{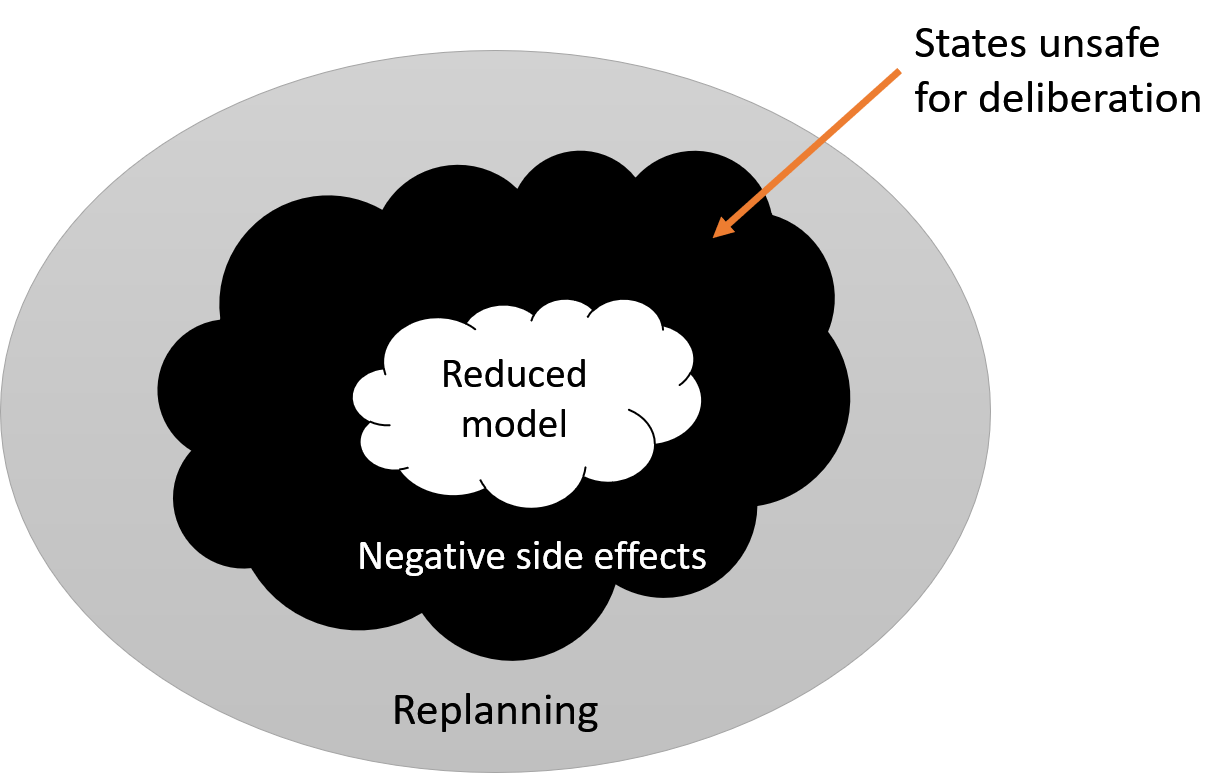}
	\caption{Reduced models are complemented by replanning when the agent encounters a state for which the policy computed using the reduced model does not specify an action. Some states may be unsafe for deliberation (replanning), which is considered a negative side effect of planning with reduced models.}
	\label{fig:illustration}
	\vspace{-2pt}
\end{figure}

\section{Introduction}
A \emph{Stochastic Shortest Path} (SSP) problem is a \emph{Markov Decision Process} (MDP) with a start state and goal state(s), where the objective is to find a policy that minimizes the expected cost of reaching a goal state from the start state~\cite{bertsekas1991analysis}. Solving large SSPs optimally and efficiently is an active research area in automated planning, with many practical applications. Given the computational complexity of solving large SSPs optimally~\cite{littman1997probabilistic}, there has been considerable interest in developing efficient approximations, such as reduced models, that trade solution quality for computational gains. Reduced models simplify the problem by partially or completely ignoring uncertainty, thereby reducing the set of reachable states for the planner. We consider reduced models in which the number of outcomes per action is reduced relative to the original model.

Due to model simplification, an agent may encounter unexpected states during plan execution, for which the policy computed using reduced models may not specify an action. In such instances, the agent needs to \emph{replan} online and find a new way to reach a goal state. We consider this as a side effect of using reduced models as the agent never has to replan when the complete model is used. Current reduced model techniques assume that the agent can replan in any state~\cite{yoon2007ff,keller2011polynomial,saisubramanianoptimizing}. That is, they are oblivious to how the ignored outcomes may affect the system. Some of these side effects may be acceptable to the user, but others may be unsafe even though they accelerate planning (Figure~\ref{fig:illustration}). 

Consider, for example, a robot navigating on a sidewalk adjacent to a street. The robot has a $0.1$ probability of sliding on to the street when navigating at high speed and $0.02$ probability of sliding to the street when navigating at a low speed. When planning with a reduced model that ignores the possibility of sliding as an outcome, the robot will have to replan if the ignored outcome occurs during plan execution. However, it is risky for the robot to replan while on the street and it is expected to be better prepared to handle such situations. Another example of replanning in risky states is an autonomous vehicle replanning in the middle of an intersection after missing a turn or an exit. In the terminology introduced by~\citeauthor{amodei2016concrete}~(\citeyear{amodei2016concrete}), these are \emph{negative side effects} of planning with reduced models | an undesired behavior that is the result of ignoring certain outcomes in planning. Table~\ref{tab:classification} outlines various properties of negative side effects. 

The importance of accounting for risks in AI systems is attracting growing interest~\cite{kulic2005safe,Zaaai15}. One such specific form of risk is the negative side effects~\cite{amodei2016concrete} that occur when an autonomous agent optimizes an objective function that focuses on certain aspects of the environment,  leading to undesirable effects on other aspects of the environment. Recently, researchers have proposed techniques to address the negative side effects that arise due to misspecified rewards~\cite{hadfield2017inverse} and incomplete model~\cite{zhang2018minimax}. In this work, we focus on addressing negative side effects in the form of deliberation in unsafe states that arise due to planning with simplified or reduced models. This is a type of irreversible negative side effect that is fully observable and avoidable. The severity of replanning may be mild or significant depending on the state. While the negative side effects can always be reduced by using the full model, this defeats the purpose of using reduced models. The challenge is to plan with reduced models while simultaneously being prepared for handling risky situations. This can be achieved by formulating reduced models that can adapt to risks by selectively improving the model fidelity. 

We propose planning using a portfolio of outcome selection principles to minimize the negative side effects of planning with reduced models. The model fidelity is improved in states that are risky | unsafe for deliberation or replanning. We consider a factored state representation in which the states are represented as a vector of features. Given a set of features that characterize states unsafe for deliberation, our approach employs more informative outcomes or the full model for actions in states leading to these side effects and a simple outcome selection otherwise. To identify states where model fidelity is to be improved, the agent performs reachability analysis using samples and the given features. The sample trajectories are generated by depth-limited random walk. Reachability analysis is performed primarily with respect to negative side effects and this is used as a heuristic for choosing the outcome selection principles in a portfolio. We evaluate our approach in three proof-of-concept domains: an electric vehicle charging problem using real-world data, a racetrack domain and a sailing domain. Our empirical results show that our approach significantly minimizes the negative side effects, without affecting the run time. While we address one particular form of negative side effects in planning using reduced models, our proposed framework is generic and can be extended to potentially address other forms of negative side effects.

Our primary contributions are: (i) formalizing negative side effects in planning using reduced models; (ii) introducing a portfolio of reduced models approach to minimize the negative side effects; and (iii) empirical evaluation of the proposed approach on three domains. 

\begin{table} 
	\centering 
	\renewcommand{\arraystretch}{1.2}
	\begin{tabular}{|l|l|}
		\hline
		Property  &  Property Values
		\\ \hline \hline
		Impact & Reversible or irreversible \\
		Severity & Significant or insignificant \\
		Occurrence & Avoidable or unavoidable \\
		Observability & Fully observable or partially observable \\
		 \hline
	\end{tabular}
	\caption{Classification of negative side effects.}
	\label{tab:classification}
\end{table}

\section{Background}
This section provides a brief background on stochastic shortest path problems and planning using reduced models.

\subsection{Stochastic Shortest Path Problems}
A Stochastic Shortest Path (SSP) MDP defined by $M= \langle S,A,T, C,s_0,S_G \rangle$, where 
\begin{itemize}
	\item $S$ is a finite set of states;
	\item $A$ is a finite set of actions;
	\item $T:S\times A\times S \rightarrow [0,1]$ is the transition function which specifies the probability of reaching a state $s'$ by executing an action $a$ in state $s$ and is denoted by $T(s,a,s')$;
	\item $C:S\times A \rightarrow \mathbb{R^+}\cup\{{0}\}$ specifies the cost of executing an action $a$ in state $s$ and is denoted by $C(s,a)$;
	\item $s_0 \in S$ is the start state; and 
	\item $S_G \subseteq S$ is the set of absorbing goal states.
\end{itemize}
The cost of an action is positive in all states except goal states, where it is zero. The objective in an SSP is to minimize the expected cost of reaching a goal state from the start state. It is assumed that there exists at least one proper policy, one that reaches a goal state from any state $s$ with probability 1.
The optimal policy, $\pi ^*$, can be extracted using the value function defined over the states, $V^*(s)$: 
\begin{align} 
V^*(s) &= \min_a ~~Q^*(s,a), \hspace{10pt} \forall s \in S \label{V-value} \\ 
Q^*(s,a)& =  C(s,a) \!+\! \sum_{s'}T(s,a,s') V^*(s'), \forall (s, a)
\label{Q-value} 
\end{align} 
where $Q^*(s,a)$ denotes the optimal Q-value of the action $a$ in state $s$ in the SSP $M$~\cite{bertsekas1991analysis}. 

\subsection{Planning Using Reduced Models}
While SSPs can be solved in polynomial time in the number of states, many problems of interest have a state-space whose size is exponential in the number of variables describing the problem~\cite{littman1997probabilistic}. This complexity has led to the use of approximate methods such as reduced models. Reduced models simplify planning by considering a subset of outcomes. Let $\theta(s,a)$ denote the set of all outcomes of $(s,a)$, $\theta(s,a)=\{s' \vert T(s,a,s')\!>\!0\}$.

\begin{definition}
A \textbf{reduced model} of an SSP $M$ is represented by the tuple $M' \!=\!\langle S,A,T', C, s_0,S_G \rangle$ and characterized by an altered transition function $T'$ such that $\forall (s,a) \!\in \! S \times A, \theta'(s,a) \subseteq \theta(s,a)$, where $\theta'(s,a) \!=\! \{s' \vert T'(s,a,s')\!>\!0\}$ denotes the set of outcomes in the reduced model for action $a\in A$ in state $s \in S$. 
\end{definition}
We normalize the probabilities of the outcomes included in the reduced model, but more complex ways to redistribute the probabilities of ignored outcomes may be considered. The outcome selection process in a reduced model framework determines the number of outcomes and how the specific outcomes are selected. Depending on these two aspects, a spectrum of reductions exist with varying levels of probabilistic complexity.
	
The \emph{outcome selection principle} (OSP) determines the outcomes included in the reduced model per state-action pair, and the altered transition function for each state-action pair. The outcome selection may be performed using a simple function such as always choosing the most-likely outcome or a more complex function. Typically, a reduced model is characterized by a single OSP. That is, a single principle is used to determine the number of outcomes and how the outcomes are selected across the entire model. Hence, existing reduced models are incapable of selectively improving model fidelity.

\section{Minimizing Negative Side Effects}
One form of negative side effects of planning using reduced models is replanning in states, as a result of ignoring outcomes, which are prescribed by a human as unsafe for deliberation. The set of states characterizing negative side effects for planning with a reduced model $M'$, denoted by $NSE(M')$, is defined as $NSE(M') = \{s' \vert T(s,a,s')>0 \land D(s') = true \land s' \notin \theta'(s,a)\}$, where $D(s')$ is true if the deliberation in the state is \emph{risky} during policy execution. Thus, the OSP determines the reduced model and thereby the resulting negative side effects. The challenge is to determine outcome selection principles that minimize the negative side effects of planning with reduced models without significantly compromising the run time gains.

While the negative side effects can always be minimized using the full model or picking the right determinization, the former is computationally complex and finding the right determinization is a non-trivial meta-level decision problem~\cite{PZicaps14}. Another approach is to penalize the agent for replanning in risky states; that is, not accounting for the risky outcomes in the reduced model formulation. However, this is useful only when the agent performs an exhaustive search in the space of reduced models, which is computationally complex. Given a set of features characterizing states unsuitable for deliberation, $D(\vec{f})$, we address this issue by providing efficient reduced model formulation that can selectively improve model fidelity. A straightforward approach is to improve the model fidelity in risky states. However, this does not minimize the negative side effects and it is more beneficial to improve the model fidelity in states leading to these risky states. Based on this insight, we present a heuristic approach, which is based on reachability to these risky states, to select outcome selection principles for each state-action pair.

\subsection{Portfolios of Reduced Models}
We present \emph{planning using a portfolio of reduced models} (PRM), a generalized approach to formulate reduced models, by switching between different outcome selection principles, each of which represents a different reduced model. The approach is inspired by the benefits of using portfolios of algorithms to solve complex computational problems~\cite{PZamai06}.

\begin{definition}
	Given a portfolio of finite outcome selection principles, $\mathrm{Z} = \{\rho _1, \rho_2,...,\rho_k\}$, $k\!\!>\!\!1$, a \textbf{model selector}, $\Phi$, generates $T'$ for a reduced model by mapping every $(s,a)$ to an outcome selection principle, $\Phi\!: S\times A \rightarrow \rho_i$, $\rho_i \in \mathrm{Z}$, such that $T'(s,a,s') = T_{\Phi(s,a)}(s,a,s')$, where $T_{\Phi(s,a)}(s,a,s')$ denotes the transition probability corresponding to the outcome selection principle selected by the model selector.
\end{definition}
Clearly, the existing reduced models are a special case of PRMs, with a model selector ($\Phi$) that always selects the same OSP for every state-action pair. In planning using a portfolio of reduced models, however, the model selector typically utilizes the synergy of multiple outcome selection principles. Each state-action pair may have a different number of outcomes and a different mechanism to select the specific outcomes. Hence, we leverage this flexibility in outcome selection to minimize negative side effects in reduced models by using more informative outcomes in the risky states and using simple outcome selection principles otherwise. Though the model selector may use multiple outcome selection principles to generate $T'$ in a PRM, note that the resulting model is still an SSP. In the rest of the paper, we focus on a basic instantiation of a PRM that alternates between the extremes of reductions. 

\begin{definition} A \textbf{0/1 reduced model} (\zo) is a PRM with a model selector that selects either one or all outcomes of an action in a state to be included in the reduced model. 
\end{definition}
A \zo~is characterized by a model selector, $\Phi_{\mbox{\tiny 0/1}}$, that either ignores the stochasticity completely ($0$) by considering only one outcome of $(s,a)$, or fully accounts for the stochasticity ($1$) by considering all outcomes of the state-action pair in the reduced model. For example, it may use the full model in states prone to negative side effects, and determinization otherwise. Thus, a \zo~that guarantees goal reachability with probability 1 can be devised, if a proper policy exists in the SSP. Our experiments using {\zo} show that even such basic instantiation of a PRM works well in practice. 

Devising an efficient model selector automatically can be treated as a meta-level decision problem that is computationally more complex than solving the reduced model, due to the numerous possible combinations of outcome selection principles. Even in a {\zo}, finding when to use determinization and when to use the full model is non-trivial.  In the more general setting, all OSPs in $\mathrm{Z}$ may have to be evaluated to determine the best reduced model formulation in the worst case. Let $\tau_{max}$ denote the maximum time taken for this evaluation across all states. In a PRM with a large portfolio, the OSPs may be redundant in terms of specific outcomes. For example, selecting the most-likely outcome and greedily selecting an outcome based on heuristic could result in the same outcome for some $(s,a)$ pair. We use this property to show that the worst case time complexity of devising an efficient model selector is independent of the size of the portfolio, which may be much larger than the size of the problem under consideration, and depends only the size of states and actions. The following proposition formally proves this complexity.

\begin{proposition}
	The worst case time complexity for $\Phi_{\mbox{\tiny 0/1}}$ to generate $T'$ for a \zo~is $\mathcal{O}(\vert A\vert  \vert S \vert^{2} \tau_{max}).$
\end{proposition}
\begin{proofsketch} For each $(s,a)$, at most $\vert \mathrm{Z} \vert$ outcome selection principles are to be evaluated and this takes at most $\tau_{max}$ time. Since this process is repeated for every $(s,a)$, $\Phi$ takes $\mathcal{O}(\vert S\vert \vert A\vert \vert \mathrm{Z} \vert \tau_{max})$ to generate $T'$. In the worst case, every action may transition to all states and the outcome selection principles in $\mathrm{Z}$ may be redundant in terms of the number and specific outcomes set produced by them. To formulate a $0/1$ RM of an SSP, it may be necessary to evaluate every outcome selection principle that corresponds to a determinization or a full model. The set of unique outcomes, $k$, for a $0/1$ RM is composed of all unique deterministic outcomes, which is every state in the SSP, and the full model, $\vert k \vert \leq \vert S \vert +1$. Replacing $\vert Z \vert$ with $\vert k \vert$, the complexity is reduced to $\mathcal{O}(\vert A\vert \vert S \vert^{2} \tau_{max})$.	 
\end{proofsketch}

\begin{figure}
	\subfigure[Orignal Problem]{\includegraphics[scale=0.44]{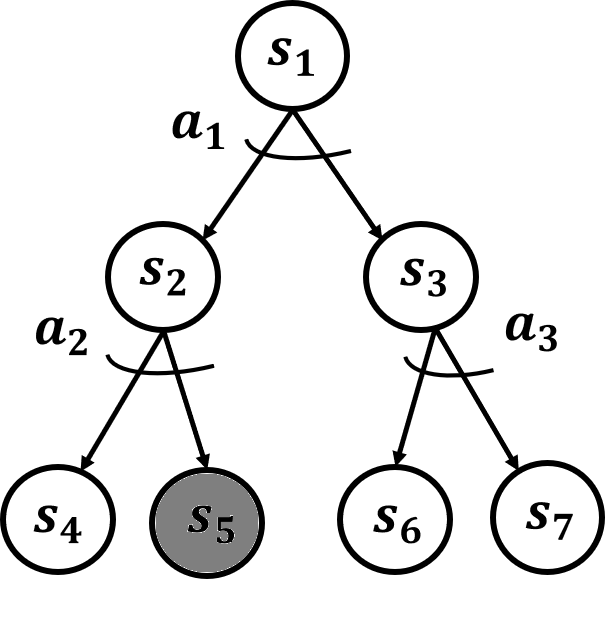}} \hfill
	\subfigure[MLOD]{\includegraphics[scale=0.43]{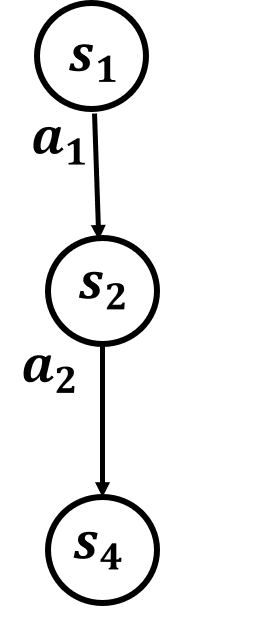}} \hfill
	\subfigure[0/1 RM]{\includegraphics[scale=0.4]{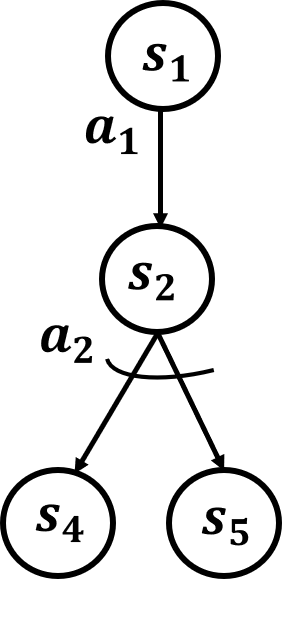}}
	\caption{Examples of most likely outcome determinization (MLOD) and 0/1 RM of a problem. The shaded state $s_5$ in the original problem denotes a risky state. }
	\label{fig:example}
\end{figure}

\subsection{Heuristic Approach to Model Selection}
Since it is computationally complex to perform an exhaustive search in the space of possible {\zo} models, we present an approximation to identify when to use the full model, using samples.

Given the features characterizing negative side effects, model selection is performed by generating and solving samples. The samples are generated automatically by sampling states from the target problem. Alternatively, small problem instances in the target domain may be used as samples. In this paper, smaller problems are created automatically by multiple trials of depth limited random walk on the target problems and solved using LAO*~\cite{hansen2001lao}. The probability of reaching the states unsafe for deliberation is computed for these samples, based on the features. Trivially, as the number of samples and the depth of the random walk are increased, the estimates converge to their true values. The full model is used in states that meet the reachability threshold. In all other states, determinization is used. 

Figure~\ref{fig:example} is an example of an SSP with a risky state (shaded) and instantiations of reduced models formed with uniform and adaptive outcome selection principles. The most likely outcome determinization (MLOD) uses uniform outcome selection principle -- always selects the most likely outcome -- and may ignore $s_5$ in the reduced model. A {\zo}, however, uses the full model in state action pair leading to the risky state, minimizing negative side effects and resulting in a model that is better prepared to handle the risky state. The reachability threshold can be varied to devise reduced models with different fidelity with varying proportions of full model usage, resulting in different sensitivity to negative side effects and computational complexity. 

\section{Experimental Results}
We experiment with the {\zo} on three proof-of-concept domains including an electric vehicle charging problem using real world data from a university campus, and two benchmark planning problems: the racetrack domain and the sailing domain. The experiments employ a simple portfolio $Z = \{\text{most-likely outcome determinization\,(MLOD), full model}\}$. We compare the results of {\zo} to that of MLOD and M02 reduction,  which greedily selects two outcomes per each state-action pair. 

The expected number of negative side effects, cost of reaching the goal, and planning time are used as evaluation metrics. For {\zo}, we use the full model in states that had a reachability of 0.25 or more to at least one of the risky states, which is estimated using thirty samples. In all other states, MLOD is used. All results are averaged over 100 trials of planning and execution simulations and the average times include re-planning time. The deterministic problems are solved using the A* algorithm~\cite{hart1968formal} and other problems using LAO*. All algorithms were implemented with $\epsilon{=}10^{-3}$ and using $h_{min}$ heuristic computed using a labeled version of LRTA*~\cite{korf1990real}.  

\subsection{EV Charging Problem}
The electric vehicle (EV) charging domain considers an EV operating in a vehicle-to-grid (V2G) setting in which an EV can either charge or discharge energy from the smart grid~\cite{ma2012modeling,saisubramanianoptimizing}. By planning when to buy or sell electricity, an EV can devise a robust policy for charging and discharging that is consistent with the owner's preferences, while minimizing the long-term cost of operating the vehicle. 

We consider uncertainty in parking duration, which is specified by a probability that certain states may become terminal states. This problem is modeled as a discrete finite-horizon MDP with maximum parking duration as the horizon $H$. Each state is represented by the tuple $\langle l, t, d, p, e \rangle$, where $l$ denotes the current level of charge of the vehicle, $t \leq H$ denotes the current time step. $d $ and $p$ denote the current demand level and price distribution for electricity and $0\!\leq\!e\!\leq 3 $ denotes the anticipated departure time specified by the owner. If the owner has not provided this information, then $e\!=\!3$ and the agent plans for $H$. Otherwise, $e$ denotes the time steps remaining for departure. The vehicle can charge $(\Omega_i^+)$ and discharge $(\Omega_i^-)$ at three different speed levels, where $1 \leq i \leq 3$ denotes the speed level, or remain idle ($\varnothing$). The $\Omega_i^+$ and $\Omega_i^-$ actions are stochastic, while the $\varnothing$ action is deterministic. The objective is to find a reward maximizing policy $\pi ^* : S \rightarrow A$ that maximizes goal reachability, given the charging preferences of the vehicle owner. If the exit level charge (goal) requirement of the owner is not met at departure, then the owner may need to extend the parking duration to reach the required charge level. Since this is undesirable, any state from where the goal cannot be reached in the remaining parking duration is treated as a preference violation (risk) in the MDP with a large penalty.

Four demand levels and two price distributions are used in the experiments. Each $t$ is equivalent to 30 minutes in real time. We assume that the owner may depart between four to six hours of parking with a probability of $0.2$ that they announce their planned departure time. Outside that window, there is a lower probability of $0.05$ that they announce their departure. The charging rewards and the peak hours are based on real data~\cite{pricing}. The battery capacity and the charge speeds for the EV are based on Nissan Leaf configuration. The charge and discharge speeds are considered to be equal. The battery inefficiency is accounted for by adding a $15\%$ penalty on the rewards. The negative side effects are estimated using: time remaining for departure, if the current time is peak hour, and if there is sufficient charge for discharging, with one step-lookahead.

\vspace{5pt}
\noindent{\textbf{EV Dataset} } 
The data used in our experiments consist of charging schedules of electric cars over a four month duration in 2017 from an American university campus. The data is clustered based on the entry and exit charges, and we selected 25 representative problem instances across clusters for our experiments. The data is from a typical charging station, where the EV is unplugged once the desired charge level is reached. Since we are considering an extended parking scenario as in a vehicle parked at work, we assume parking duration of up to eight hours in our experiments. Therefore, for each instance, we alter the parking duration.
\subsection{Racetrack Domain}
We experimented with six problem instances from the racetrack domain~\cite{barto1995learning}, with modifications to increase the difficulty of the problem. We modified the problem such that, in addition to a 0.10 probability of slipping, there is a 0.20 probability of randomly changing the intended acceleration by one unit. The negative side effects are estimated using one-step lookahead and state features such as: whether the successor is a wall or pothole or goal, and if the successor is moving away from the goal, estimated using the heuristic. 

\begin{table} 
	\centering 
	\renewcommand{\arraystretch}{1.2}
	\begin{tabular}{|l|r|r|r|}
		\hline
		Problem Instance  &  MLOD  & M02  & 0/1 RM
		\\ \hline \hline
		EV-RF-1 & 12.91 & 7.16 & 0  \\
		EV-RF-2 & 11.88 & 6.60 & 0  \\ 
		EV-RF-3 & 19.42 & 8.24 & 0  \\
		EV-RF-4 & 19.90 & 12.09 & 0 \\\hline
		Racetrack-Square-3 &20.80  & 0 & 0.20 \\
		Racetrack-Square-4 & 21.90 & 10.73 & 7.20  \\
		Racetrack-Square-5~~ & 21.20 & 16.30 & 9.03 \\
		Racetrack-Ring-3 & 23.40 & 0 & 0 \\
		Racetrack-Ring-5 & 26.20 & 11.58 & 17.00   \\
		Racetrack-Ring-6 & 28.00 & 3.33 & 3.28  \\
		\hline
		Sailing-20(C) & 51.4 & 30.26 & 0  \\
		Sailing-40(C) & 13.51 & 15.98 & 9.88 \\
		Sailing-80(C) & 50.3 & 16.7 & 13.25 \\
		Sailing-20(M) & 50.09 & 14.82 & 0.12 \\
		Sailing-40(M) & 9.59 & 33.15 & 1.11 \\
		Sailing-80(M) & 52.9 & 11.61 & 0 \\ \hline
	\end{tabular}
	\caption{Average negative side effects of different reduced models on three domains. The results are averaged over 100 trials.}
	\label{tab:replan}
\end{table}
\begin{table*}[t]  
	\centering 
	\renewcommand{\arraystretch}{1.2}
	\begin{tabular}{|l|r|r|r|r|r|r|}
		\hline
		& \multicolumn{2}{c|}{MLOD }
		& \multicolumn{2}{c|}{M02}  
		& \multicolumn{2}{c|}{0/1 RM} 
		\\ \hline
		Problem Instance  &  \multicolumn{1}{|p{1.5cm}|}{\centering \% Cost Increase} &  \multicolumn{1}{|p{1.5cm}|}{\centering \% Time Savings}  &  \multicolumn{1}{|p{1.5cm}|}{\centering \% Cost Increase} &  \multicolumn{1}{|p{1.5cm}|}{\centering \% Time Savings}  &  \multicolumn{1}{|p{1.5cm}|}{\centering \% Cost Increase} &  \multicolumn{1}{|p{1.5cm}|}{\centering \% Time Savings}  
		\\ \hline \hline
		EV-RF-1 & 28.17 & 31.03 & 48.42 & 15.96 & 7.04 & 21.47  \\
		EV-RF-2 & 35.28 & 36.29 & 35.96 & 24.93 & 5.45 & 22.17  \\ 
		EV-RF-3 & 38.55 & 35.56 & 30.61 & 29.05 & 9.59 & 28.54 \\
		EV-RF-4 & 32.29 & 55.56 & 26.85 & 29.05 & 11.11 & 46.06  \\\hline
		Racetrack-Square-3 & 24.69 & 99.73 & 22.45 & 99.45 & 0.26 & 99.59  \\
		Racetrack-Square-4 & 44.61 & 99.85 & 18.87 & 97.13 & 5.73 & 99.68  \\
		Racetrack-Square-5~~ & 33.32 & 93.83 & 8.73 & 34.21 & 5.58 & 94.34  \\
		Racetrack-Ring-4 & 8.59 & 99.20 & 14.07 & 99.13 & 2.93 & 99.06 \\
		Racetrack-Ring-5 & 58.25 & 99.55 & 30.45 & 82.91 & 17.17 & 94.35  \\
		Racetrack-Ring-6 & 30.62 & 99.26 & 6.41 & 66.05 & 8.30 & 92.07  \\
		\hline
		Sailing-20(C) & 76.79 & 92.86 & 31.83 & 49.68 & 7.11 & 85.72 \\
		Sailing-40(C) & 63.34 & 75.91 & 19.52 & 62.21 & 2.85 & 71.66 \\
		Sailing-80(C) & 39.19 & 86.99 & 13.25 & 70.27 & 2.81 & 83.88  \\
		Sailing-20(M) & 73.81 & 99.84 & 79.62 & 39.56 & 1.63 & 93.25 \\
		Sailing-40(M) & 65.64 & 94.56 & 37.75 & 88.86 & 3.08 & 85.01 \\
		Sailing-80(M) & 87.66 & 93.48 & 24.59 & 90.37 & 4.99 & 96.67 \\ \hline
	\end{tabular}
	\caption{Comparison of the solution quality of different reduced models (formed using different model selectors) on three domains. The reported values are averaged over 100 trials. }
	\label{tab:results}
	\vspace{-4pt}
\end{table*}
\subsection{Sailing Domain}
Finally, we present results on six instances of the sailing domain~\cite{kocsis2006bandit}. The problems vary in terms of grid size and the goal position (opposite corner (C) or middle (M) of the grid). In this domain, the actions are deterministic and uncertainty is caused by stochastic changes in the direction of the wind. The negative side effects are estimated using one-step lookahead and based on state features such as: the difference between the action's intended direction of movement and the wind's direction, and if the successor is moving away from goal, estimated using the heuristic value.

\subsection{Discussion}
Table~\ref{tab:replan} shows the average number of negative side effects (risky states encountered without an action to execute) of the three techniques in three domains. For the EV domain, the results are aggregated over 25 problem instances for each reward function. In our experiments, we observe that as the model fidelity is improved, the negative side effects are minimized. In all four reward functions cases of the EV, {\zo} has no negative side effects. Overall, {\zo} has the least negative side effects in all three proof-of-concept domains. 

Table~\ref{tab:results} shows the average increase in cost (\%) (with respect to the optimal cost as lower bound) and the time savings (\%) (with respect to solving the original problem). The \emph{least cost increase \%} indicate that the performance of the technique is closer to optimal. The \emph{higher time savings values} indicate improved run time gains by using the model. The runtime of {\zo} is considerably faster than solving the original problem, while yielding almost optimal solutions. This shows that our approach does not significantly affect the run time gains so as to improve the solution quality and minimize negative side effects. Note that we solve {\zo} using an optimal algorithm, LAO*, to demonstrate the effectiveness of our framework by comparing the optimal solutions of the models. In practice, any SSP solver (optimal or not) may be used to improve run time gains. Overall, {\zo} yields almost optimal results, in terms of cost and negative side effects.

\section{Related Work}
Interest in using reduced models to accelerate planning increased after the success of FF-Replan~\cite{yoon2007ff}, which won the 2004 IPPC using the Fast Forward (FF) technique to generate deterministic plans~\cite{hoffmann2001ff}. Following  the success of FF-Replan, researchers have proposed various methods to improve determinization~\cite{yoon2010improving,keller2011polynomial,issakkimuthu2015hindsight}. Robust FF (RFF) generates a plan for an envelope of states such that the probability of reaching a state outside the envelope is below some predefined threshold~\cite{teichteil2010incremental} and replans to the subset of states for which the policy is already computed. Our work differs from that of RFF since we compute reachability as a pre-processing to formulate {\zo} and not the policy. 

Safety issues in SSPs have been studied mostly with respect to unavoidable dead ends that affect the goal reachability~\cite{kolobov2012planning,kolobov2012theory,trevizan2017efficient}. Recently, researchers have focused on formulating richer reduced models that can balance the trade-off between complexity and model fidelity so as to improve the solution quality and safety when planning with reduced models~\cite{PZicaps14,saisubramanian2018planning}. Our work aims to strike a balance between minimizing negative side effects and using a simpler model for planning, bearing similarity to the works that balance the trade-off between solution quality and model fidelity in reduced model. However, unlike the existing works, we target a well-defined and a specific form of risk that arise due to replanning in a reduced model setting.

The importance of addressing negative side effects in AI systems is gaining growing interest. ~\citeauthor{amodei2016concrete}~(\citeyear{amodei2016concrete}) address the problem of negative side effects in a reinforcement learning setting by penalizing the agent, while optimizing the reward. ~\citeauthor{zhang2018minimax}~(\citeyear{zhang2018minimax}) study a form of negative side effects in MDPs that occur when an agent's actions alter the state features in a way that may negatively surprise the user. In their work, the agent queries a user to learn the features that are acceptable for modification and computes a plan accordingly. \citeauthor{hadfield2017inverse}~(\citeyear{hadfield2017inverse}) address negative side effects that arise from misspecified objectives and misspecified reward by using the specified reward as an observation to learn the true reward function.

\section{Conclusion and Future Work}
Reduced models are a promising approach to quickly solve large SSPs. However, the existing techniques are oblivious to the risky outcomes in the original problem when formulating a reduced model. We propose planning using a portfolio of reduced models that provides flexibility in outcome selection to minimize the negative side effects while still accelerating planning. We also present a heuristic approach to determine the outcome selection principles. Our empirical results demonstrate the promise of this framework; {\zo} that is based on the proposed heuristic yields improves the solution quality, apart from minimizing the negative side effects. 
Our results also contribute to a better understanding of how disparate reduce model techniques help improve the solution quality, while minimizing the negative side effects.

This paper focuses on addressing one specific form of negative side effect in planning that occurs when using reduced models. There are a number of interesting future research directions in this area. First, we are looking into ways to automatically identify risks in the system. One approach is to generate samples for learning risks using machine learning techniques such as regression and decision stump. Another approach is to employ limited user data in the form of human demonstrations to learn the feature and states that are risky for replanning. Second, we aim to build a classifier to distinguish side effects from negative side effects, which is a critical component for automatically identifying the risks. Finally, the $0/1$ RM represents an initial exploration of a broad spectrum of PRMs, which we will continue to explore and further analyze outcome selection methods for PRMs with varying levels of probabilistic complexity. 

\section{Acknowledgments}
Support for this work was provided in part by the U.S. National Science Foundation grants IIS-1524797 and IIS-IIS-1724101.

\small
\bibliographystyle{aaai}
\bibliography{References}
\end{document}